# A primer on getting neologisms from foreign languages to under-resourced languages

Luis Camacho

Pontificia Universidad Católica del Perú

camacho.l@pucp.pe

Summary: Mainly due to lack of support, most under-resourced languages have a reduced lexicon in most realms and domains of increasing importance, then their speakers need to significantly augment it. Although neologisms should arise from the languages themselves, external sources are widely accepted. However, we dispute the "common sense" of using only one foreign language, the imposed official one, which is highly probably a legacy of colonialism, and we propose to introduce neologisms from any language as long as these neologisms "sound like" native words of the target languages.

## Introduction

The word does not create the thing, but if the thing is not labeled it will be impossible for it to be identified and therefore it cannot be referred to and searched for, not even by digital means. Neologisms are certain uses, expressions, and words that did not traditionally exist in a language, but are incorporated into it due to the need of speakers to adapt to a new reality [1]. That is, neologisms are those new words and expressions that speakers incorporate into a language, as new things and new ways of doing to name arise. They are the exact opposite of archaisms. The appearance of neologisms is a common and ordinary process in all languages, forced as they are to adapt and update or die. However, a word can be considered a neologism only for a certain time, since once it has been incorporated and normalized as part of the language, it simply ceases to be a novelty.

The simplest way to classify neologisms would be from the method used to create them, thus we have:

1. morphological neologisms: they are built using words that already exist in the language, through the processes of composition or derivation. For example, the word "aircraft" was once a neologism, made up of the prefix "air" and the suffix "craft". This also happens with "teleoperators" or with "biosecurity".
2. semantic neologisms: on the other hand, they are obtained when the same word that already exists in the language acquires new meanings, more or less linked to those it already had. This is what happened with the word "virus" since the appearance of malicious software on the Internet, or with "browsing" since the possibility of entering the web: both terms already existed, but were applied to other areas.
3. foreign words. when they come from other languages, whether or not we respect their form and pronunciation.

Neologisms are still considered to be introduced one by one in communities according to need. However, strictly speaking, it hasn't been that way for centuries. The massification of school education, which began two centuries ago, suddenly introduced not a few but many unsolicited neologisms that little by little were incorporated into popular speech transmitted by school children to parents with little or no literacy, we are talking about words that were perhaps already present in the languages of instruction but were unknown to the masses [2]. In the same direction, radio and television networks, which became massive less than a century ago, quickly became a constant second source of introducing new words to the masses, whether they were completely new (true neologisms) or slang of one or several groups. Finally, in this century the Internet is taking charge of popularizing words that are coined daily in any part of the connected world [3]. The point is that *stricto sensu*, contrary to what is still held, neologisms are not created under the demand of small groups, but rather the "global village" creates many neologisms almost daily and launches them at its global audience, who are exposed to them even if they have not requested them.

We focus on the automation of searching and proposition of neologisms for the Quechua language coming from any other language considering the proposed words must match the phonology of the target. The Quechua language is the most spoken native language of America. Our contributions are:

1. code to search neologisms from any language which have a phonetic notation database
2. we present a database composed of 41722 proposed neologisms

**Phonemic orthography**

A phonemic orthography is a system for writing a language in which the graphemes correspond to the phonemes of the language. Natural languages rarely have perfectly phonemic orthographies; a high degree of grapheme-phoneme correspondence can be expected in orthographies based on alphabetic writing systems, but they differ in how complete this correspondence is. English orthography, for example, is alphabetic but highly nonphonemic; it was once mostly phonemic during the Middle English stage when the modern spellings originated, but spoken English changed rapidly while the orthography was much more stable, resulting in the modern nonphonemic situation. However, because of their relatively recent modernizations compared to English, the Romanian, Italian, Turkish, Spanish, Finnish, Czech, Latvian, Esperanto, Korean, and Swahili orthographic systems come much closer to being consistent phonemic representations [6].

In an ideal phonemic orthography, there would be a complete one-to-one correspondence (bijection) between the graphemes (letters) and the phonemes of the language, and each phoneme would invariably be represented by its corresponding grapheme. So the spelling of a word would unambiguously and transparently indicate its pronunciation, and conversely, a speaker knowing the pronunciation of a word would be able to infer its spelling without any doubt. That ideal situation is rare but exists in a few languages.

In less formal terms, a language with a highly phonemic orthography may be described as having regular spelling. Another terminology is that of deep and shallow orthographies, in which the depth of orthography is the degree to which it diverges from being truly phonemic. Xavier Marjou[7] used an artificial neural network to rank 17 orthographies according to their level of Orthographic depth. Among the tested orthographies, Chinese and French orthographies, followed by English and Russian, are the most opaque regarding writing (i.e. phonemes to graphemes direction), and English, followed by Dutch, is the most opaque regarding reading (i.e. graphemes to phonemes direction); Esperanto, Arabic, Finnish, Korean, Serbo-Croatian and Turkish are very shallow both to read and to write; Italian is shallow to read and very shallow to write, Breton, German, Portuguese and Spanish are shallow to read and to write.

**Quechua language**

This language began to be written, with Latin characters, approximately 500 years ago, after the Spaniard invasion of South America began; there is no evidence that native Quechua speakers used any kind of script before then. Since the beginning of the 20th century, there have been several attempts to provide the Quechua and Aymara languages with their own alphabets. Only recently, in October 1975, there was a serious attempt when the Ministry of Education appointed a commission to implement the law for the officialization of the Quechua language. That commission recommends the General Basic Alphabet of Quechua, which was approved by the Ministry through RM No. 4023-75-ED. The few institutions that promoted bilingual intercultural education in Peru implemented this alphabet in their respective programs, however, difficulties arose especially with the use of vowels. After 2 years of workshops led by San Marcos National University and by San Cristóbal Huamanga National University, in 1985 the Quechua pan-alphabet was made official through RM 1218-85-ED. It is a set of 34 graphemes, from which each dialect or macro dialect can select a subset according to its phonological characteristics, in addition to some general indications of orthographic and punctuation rules. Starting in 2005 and up to the present, the Ministry of Education has organized a set of participatory events in a more systematic way to implement RM 1218-85-ED throughout the country.

Quechua has a phonemic orthography instead of a phonetic orthography (or more properly, instead of an allophonic orthography). Quechua follows three criteria of unified writing: phonological, morphological and lexical richness. [8]

The southern Quechua alphabet, by far the most spoken dialect, has a total of 28 spellings, 3 vowel spellings and 25 consonant spellings, as shown in Table 1. Of all the spellings, 15 consonants and the 3 vowels represent phonemes common to the two sub-dialects: chanka and qullaw. The remaining 10 consonants represent laryngealized phonemes that are only found in the qullaw dialect. The appearance of these letters, within the words, will be determined by the restrictions of the phonotactics of this language. The most important restrictions are:

1. The next letters can't be located at the end of any word:
   h, ñ, chh, ch', kh, k', ph, p', qh, q', t, th, t'
2. Pronunciation of sounds [I] or [e] depends on the presence of the consonants /q/, /q'/ or /qh/. This means that those sounds aren't phonemes or they are not part of the Quechua phonological system, but rather, they are just allophones of the vowel /i/ that appear in a certain and predictable context.
3. Pronunciation of sound [o] depends on the presence of the consonants /q/, /q'/ or /qh/. This means that sound is not a phoneme or it is part of the Quechua phonological system, but rather, it is just an allophone of the vowel /u/ that appears in a certain and predictable context.

4. Sound /ʎ/ at the end of a syllable followed by /q/ must be written <ll> instead of <l>; following the criterion of normalized writing, where the most conserved form is used.

5. Neither two vowels nor two consonants should be placed consecutively, much less three or more.

6. In order not to confuse the use between the spellings /k/ and /q/ at the end of a syllable, phonological awareness is important; that is, to realize if the vowels /i/ and /u/ present any vowel opening in words such as: <wiksa>, <liklla>, <pikchu> o <t'iktu/tiktu>. If the opening does not increase, the spelling <k> corresponds; otherwise, the spelling <q> corresponds, as in <tiqti>, <yachachiq>, <luqlu> or <phuqchiy/puqchiy>.

7. When /m/ is found as part of the root word, it will always be written with <m> before /p/, as in "pampa" (plain) or "t'impuy/timpuy" (to boil). On the other hand, if the root word ends in /n/, eg "purun" or "allin" it will remain <n> in writing, as in "purunpi". In the same way, when the word ends in the suffix {-n}, which expresses possessive of the third person singular, even if it sounds like [m] at the phonetic level, if it precedes suffixes that begin with the sound /p/ (for example, -pi, -paq, etc), the suffix {-n} remains written <n>, as in "wamp'unpaq/ wampunpaq" ("for his ship") or "llaqtanpi" ("in his town").

Additionally, there are restrictions in the Quechua syllabic structure such as:

1. Vowels <a>, <i> y <u>, being syllabic core, form all types of syllables in the Quechua language.

| Syllable type | Example |
| --- | --- |
| V | "i" as in "iru" |
| VC | "pu" as in "ullpu" |
| CV | "ma" as in "maki" |
| CVC | "tan" as in tankar |

2. Consonants <p>, <k>, <q>, <ch>, <l>, <ll>, <m>, <n>, <r>, <s>,<t>, <w> y <y> can be located at the end of a syllable, therefore, they can form closed-type syllables such as: VC or CVC, while consonants <h> (except onomatopoeic sounds), <ñ>, and laryngealized consonants <chh>, <ch'>, <kh>, <k'>, <ph>, <p'>, <qh>, <q' >, <th> and <t'> are prohibited at the end of a syllable, therefore, with this type of consonant, neither VC nor CVC type syllables can be formed.

| Syllable type | Example |
| --- | --- |
| VC | "un" as in "unquy" |
| CVC | "wak" as in "wakcha" |

3. All consonants can form syllables of the CV type and can be the first consonant of the CVC syllable type.

| Syllable type | Example |
| --- | --- |
| CV | "ña" as in "ñaña" |
| CVC | "qay" as in "tanqay" |

**Materials and methods**

The coining of novel words relies on human creativity, with the new terms often conveying a lot of information in an inventive way. Work on the computational generation of neologisms mostly focused on creating compounds and word blends from source words. Moran Mizrahi et al. [9] went a step further, they set out to explore the possibility of automating some of this inherently human, creative linguistic process, and they focused on the Hebrew language.

We don't walk Mizrahi's route but we do use computational linguistics not to generate but to search real words from any language, those words that sound like Quechua words. This method avoids the rephonologization of the proposed neologisms since they have been selected by our algorithm precisely because they meet the requirement of fitting the phonology of the target language. Our method requires phonetic notation of the source languages. We found two sources of IPA representation of a number of languages: Open Dictionary[1] and Wikipron[2]. The Open Dictionary project aims to collect open-licensed multilingual dictionary data and provide it in a variety of accessible formats for use by humans and computers.

WikiPron[9] is a command-line tool and Python API for mining multilingual pronunciation data from Wiktionary, as well as a database of pronunciation dictionaries mined using this tool.

These two sources were processed separately. The first source is made up of csv files and the second source is made up of tsv files. Each of these files corresponds to a different language, in each file, there are two columns: the words are located in the first one and their corresponding IPA transcriptions are located in the second one, in the same row. The GettingNeologisms.ipynb[3] script collects all the spelling and pronunciation rules mentioned in the previous section. After executing the script, two intermediate files are obtained that collect all the words with the requested phonology present in the csv and tsv directories, respectively. Since the files in the csv directory are labeled according to the ISO639-1 terminology, they had to be relabeled following the correspondence table to obtain the labels with the current ISO639-5 nomenclature.

The two intermediate files were inputs to the SincronizationWordNet.ipynb[4] script which identifies the words and their respective languages present in the intermediate files; if the script finds both in WordNet it delivers the English translation. To be exact, the English synset is searched for in WordNet and then all the lemmas associated with that synset are listed.

As the next step, we translated the IPA notation of the selected neologisms to the Quechua orthography. Finally, we used the GOOGLETRANSLATE function of Google Spreadsheets to translate the neologisms from their original language to Spanish. The final databases are stored in GitHub[5].

**Results**

---

Results are shown in tables 2 and 3. The first source has 3838348 words, of which 14970 (0.39%) neologisms were found and 1768 (11.81%) were translated into English. The second source has 3506258 words, of which 26752 (0.76%) were found and then 6118 (22.87%) were translated into English.

From the first source, the contribution is made by 14 out of 18 languages, but we found translations to English in WordNet for words from only 6 languages. In the second case, it was 217 out of 245 but words from only 27 languages were translated into English.

Unfortunately, WordNet is not enough populated with data, not only has a small set of languages, 32, but also the quantity of words per language is also short, which is why the rates of translation to English were just 11.81% and 22.87% respectively. Finally, 13202 plus 20634 are eligible but they are not been translated yet into English. However, just closing this paper we discovered the option to automatize translation using Google Spreadsheets, so, we used it but only to translate to the Spanish language, considering that most people will look for this pair.

## Discussion

We discuss two topics: the pertinence and results of this research.

About pertinence, it must be said the introduction of neologisms from foreign sources is more complex in the case of endangered languages, sometimes neologisms are not accepted for more social and political than technical reasons. For example, a teacher can create some neologisms for numbering with a solid linguistic foundation; nevertheless, when diffusion is made through the written materials or through training teachers, the proposal is rejected by other speakers and by the community because it was not a consensus building of a group prestigious or representative of this town and because he did not have the collaboration of those responsible elected or selected by decision-makers of first nations. For these reasons, many authors such as Carvajal [4] have agreed that these difficulties result from the diglossic situation in which endangered languages are found. In other words, European colonialism imposed European languages at top of a hierarchy to the detriment of the native languages and the baggage of knowledge of colonized people. The end of colonialism didn't change this dynamic on those lands and these European languages continue advancing and extending their use to new spaces and the native languages keep reducing their function communicative to certain cultural contexts and semantic fields (in the worst case, stop talking) and its symbolic function deteriorates and generates in its speakers questioning their identity. This limited functionality of the native languages has resulted in the gradual stagnation of the word formation mechanisms (for example, derivation and composition) and in no planning and developing neologisms for pedagogical terminology in these languages [5].

Thinking hard about this situation led to the conclusion that without diglossia, neologisms from foreign languages would be prone to be accepted. In other words, in this specific case, the problem is not the foreign source of the neologisms but the imposition of neologisms from the Spanish language, a legacy of colonialism. In that direction, we must remember that Esperanto was created artificially with the intention of serving as a *lingua franca* for international communication avoiding language dominance,  but fewer than 200 000 people speak it today, the initiative never got momentum probably because of its non-natural origin [10]. But the idea is right, our proposal reflects the same spirit: the creation of neologisms from foreign languages but not only from the neighboring languages or from the dominant languages but from any language using computational methods to select these words from any language which could fit in the target languages. This method, in the best-case scenario, would create not a *lingua franca* but a set of common words for many languages.

But, we must emphasize that this is a proposal, an extension of the universe of possibilities to introduce neologisms, we fully agree that the first source of neologisms should be the target language itself. The next step will be to send the proposal not only to specialists but to the entire Quechua-speaking population through a webpage. The digital tool most used by the Quechua-speaking community is the Qichwa2.0[6] webpage, which is a search engine that has the largest number of Quechua-Spanish bilingual dictionaries indexed. Our database will be added there as a dictionary of neologisms, putting it very clear that it is not an authoritative source of translation but a proposal in search of exposure and consensus. We must point out another virtue of this proposal: the realization that Quechua is a language and is not a dialect of Spanish, so it can establish direct relations with any other language without the mediation of the Spanish language, and here we are building the bridges that allow direct contact between the speakers of the Quechua language and the speakers of any other one.

Regards to results, all the chosen words are morphologically correct which proves that the rules used to encode the selection algorithm are correct. The percentage of words selected over words examined does not exceed 1%, this is a big surprise because we expected a much greater number. The explanation for this is the precision level of the sources, there is a distinction between broad transcription and narrow transcription. Broad transcription indicates only the most noticeable phonetic features of an utterance, whereas narrow transcription encodes more information about the phonetic characteristics of the allophones in the utterance. The advantage of narrow transcription is that it can help learners to produce exactly the right sound and allows linguists to make detailed analyses of language variation [12]. The disadvantage is that a narrow transcription is rarely representative of all speakers of a language. A further disadvantage of narrow transcription is that it involves a larger number of symbols and diacritics that may be unfamiliar to non-specialists. The advantage of broad transcription is that it usually allows statements to be made which apply across a more diverse language community. It is thus more appropriate for the pronunciation data in foreign language dictionaries, which may discuss phonetic details in the preface but rarely give them for each entry. A rule of thumb in many linguistics contexts is to use a narrow transcription when necessary for the point being made, but a broad transcription whenever possible. Our sources have both kinds of notations but we didn't code diacritics, and that is why we didn't get more eligible words.

---

[6] https://www.dic.qichwa.net/#/

**Future work**

Despite the fact we examine seven million words, this is a small fraction of the universe of words in all languages, the challenge is to generate phonetic notation of all words. In that direction is promising the research done by Marjou [11] to automatically extract the IPA phonemic pronunciation of a word based on its audio pronunciation. Since get that goal may take time, in the meantime, the languages must be classified according to their spelling and start with those whose spelling is regular and don't need the IPA notation, the paradigmatic case is the pinyin.

Authoritative translation sources and translation of words to the English language are also needed. In this first experience, only the words recorded in Wordnet were used, perhaps using Babelnet, the number can be significantly increased without discarding *googletranslate* (function of Google spreadsheets).

The low percentages of eligible words suggest that it is necessary to adjust the requirements, perhaps allowing other vocal sounds (theoretically, Quechua has only 11 of the 33 vowels)  and overall, coding diacritics.

Thinking about scalation, it is needed to normalize the code so that the same tasks can be done for any other language.

**Conclusion**

Borrowings words are a big source for enriching vocabulary. Until the past century, inadequacy and adequacy of borrowings were related to the historical development of language but in this century Internet changed the game and it became a source of daily new words for its global audience. We presented here a novel approach for coining neologisms from foreign languages to an endangered language, the Quechua. Moreover, we introduce a new paradigm: borrowing words not only from Spanish (the dominant language in South America) but from any other language whose words meet the Quechua phonology.

Neologisms should contribute to effective communication and strengthen the identity of the speakers. We agree that consensual and socially accepted neologisms are those that remain over time and enrich the vocabulary of a language; then, agreement upon by consensus among stakeholders is a must to get social acceptance. Of course, our approach is a proposal and is not an imposition.

Language dominance depends on socio-political power. With the emergence of Asian economies, Mandarin or Hindi, which together have more speakers than English, could become the next century's *lingua franca*. However, this will depend on geopolitical balances, with hyperglobalisation and deglobalisation at its extremes.

**Data Availability Statement**

The data that support the findings of this study are available in repositories GlosaCSV.xlsx and GlosaTSV.xlsx at https://github.com/luiscamachocaballero/QuechuaNeologisms. These data were derived from Open Dictionary and Wikipron, resources available in the public domain:
https://github.com/open-dict-data/ipa-dict
https://github.com/CUNY-CL/wikipron

**Acknowledgments**

This paper was written in the framework of the project Tarpuriq, linguistic resources for computational processing of Quechua language, CAP 2021 – PI0754 funded by PUCP, Catholic University of Peru

## Table 1. Official Quechua alphabet ([8] page 45)

| IPA | Quechua alphabet | Examples | English approximation |
|---|---|---|---|
| a , ɑ | a | karu | father |
| i , ɪ | i | nina | lip |
| ɛ , e | i (near uvular consonants) | qichwa, qullqi, yachachiq | pet |
| ʊ , u , ʊ | u | kusi | hook |
| ɔ , o | u (near uvular consonants) | qusqu, q'umir, atuq | more |
| č, š | ch | chunka, pacha | church |
| čʰ | chh | chhalla, chhulla | church |
| č' | ch' | ch'aska | like church but with a restriction of air |
| h | h | huñuy, muhu | house |
| k , x , k̄ | k | kuntur, puka | scat |
| kʰ | kh | khuyay | cat |
| k' | k' | k'ispa | like scar but with a restriction of air |
| l | l | layu, phalay | lamp |
| ʎ | ll | llapa, allqu | billion |
| m | m | marka, pampa | map |
| n | n | nina | none |
| ɲ | ñ | wiñay | similar to canyon |
| p , β , ɸ | p | pirqa | spat |
| pʰ , pɸ | ph | phiña, phatay | pan, sphere |
| p' | p' | p'unchaw | like spat but with a restriction of air |
| q , ɢ , χ | q | qam, nuqa | similar to scat but deeper in the throat |
| qʰ | qh | qhapaq | similar to cat but deeper in the throat |
| q' | q' | q'isa | similar to the /q/ sound but with a restriction of air |
| ɾ | r | runtu, yawar | better in American English |
| s | s | sunqu, wasi | son |
| t | t | tinku, pata | stunt |
| tʰ | th | thamay, thanta | top |
| t' | t' | t'utura, t'uqyay | like top but with a restriction of air |
| w | w | wañuy, kawsay | water |
| j | y | yana, ayni | yes |

## Table 2. Neologisms obtained from Open Dictionary

| Language | Words Total | Words Elegibles | Words Translated | Elegibles/ Total | Translated/ Elegibles |
|---|---|---|---|---|---|
| Arabic (Modern Standard) | 857161 | 23 | | 0.00% | 0.00% |
| German | 278915 | 28 | | 0.01% | 0.00% |
| English (Received Pronunciation) | 65118 | | | 0.00% | 0.00% |
| English (General American) | 125927 | 3 | 1 | 0.00% | 33.33% |
| Esperanto | 23517 | 155 | | 0.66% | 0.00% |
| Spanish (Spain) | 595896 | 2903 | | 0.49% | 0.00% |
| Spanish (Mexico) | 595885 | 2903 | 626 | 0.49% | 21.56% |
| Persian | 8090 | 60 | | 0.74% | 0.00% |
| Finnish | 92837 | 21 | 9 | 0.02% | 42.86% |
| French (France) | 245178 | 1969 | 740 | 0.80% | 37.58% |
| French (Québec) | 245958 | 2526 | | 1.03% | 0.00% |
| Japanese | 221421 | 1005 | 375 | 0.45% | 37.31% |
| Jamaican Creole | 1870 | 73 | | 3.90% | 0.00% |
| Malay (Malaysian and Indonesian) | 28215 | 693 | | 2.46% | 0.00% |
| Norwegian Bokmål | 10172 | 38 | 17 | 0.37% | 44.74% |
| Odia | 6226 | 23 | | 0.37% | 0.00% |
| Swedish | 21106 | 3 | | 0.01% | 0.00% |
| Swahili | 48308 | 2544 | | 5.27% | 0.00% |
| Vietnamese (Central) | 72234 | | | 0.00% | 0.00% |
| Vietnamese (Northern) | 72234 | | | 0.00% | 0.00% |
| Vietnamese (Southern) | 72234 | | | 0.00% | 0.00% |
| Cantonese | 56826 | | | 0.00% | 0.00% |
| Mandarin simplified | 44780 | | | 0.00% | 0.00% |
| Mandarin traditional | 48240 | | | 0.00% | 0.00% |
| **TOTAL** | **3838348** | **14970** | **1768** | 0.39% | 11.81% |

**Table 3. Neologisms obtained from Wikipron**

| Language | Words Total | Elegible Words | Words Translated | Elegible/ Total | Translated/ Elegible |
|---|---|---|---|---|---|
| Spanish; Castilian | 347921 | 5098 | 1302 | 1.47% | 25.54% |
| English | 262090 | 2209 | 1415 | 0.84% | 64.06% |
| French | 124377 | 1720 | 657 | 1.38% | 38.20% |
| Armenian | 58317 | 1322 | 0 | 2.27% | 0.00% |
| Portuguese | 36645 | 1066 | 631 | 2.91% | 59.19% |
| Ancient Greek (to 1453) | 98085 | 1029 | 0 | 1.05% | 0.00% |
| Latin | 177872 | 1015 | 0 | 0.57% | 0.00% |
| Finnish | 160781 | 800 | 422 | 0.50% | 52.75% |
| Polish | 89157 | 773 | 175 | 0.87% | 22.64% |
| Catalan; Valencian | 66099 | 765 | 470 | 1.16% | 61.44% |
| Macedonian | 52454 | 562 | 0 | 1.07% | 0.00% |
| Modern Greek (1453-) | 24032 | 556 | 93 | 2.31% | 16.73% |
| Tagalog | 16359 | 544 | 0 | 3.33% | 0.00% |
| Italian | 105951 | 502 | 180 | 0.47% | 35.86% |
| Welsh | 45653 | 470 | 0 | 1.03% | 0.00% |
| Czech | 32196 | 390 | 0 | 1.21% | 0.00% |
| Dutch; Flemish | 67862 | 328 | 190 | 0.48% | 57.93% |
| Indonesian | 7144 | 327 | 222 | 4.58% | 67.89% |
| Russian | 404598 | 285 | 0 | 0.07% | 0.00% |
| Romanian; Moldavian; Moldovan | 17115 | 284 | 119 | 1.66% | 41.90% |
| Malay (macrolanguage) | 4665 | 263 | 0 | 5.64% | 0.00% |
| German | 90873 | 258 | 0 | 0.28% | 0.00% |
| Galician | 7778 | 230 | 75 | 2.96% | 32.61% |
| other 195 languages | 1208234 | 5956 | 167 | 0.49% | 2.80% |
| **TOTAL** | 3506258 | 26752 | 6118 | 0.76% | 22.87% |